\documentclass[conference]{IEEEtran}
\IEEEoverridecommandlockouts

\usepackage{cite}
\usepackage{amsmath,amssymb,amsfonts}
\usepackage{algorithmic}
\usepackage{graphicx}
\usepackage{textcomp}
\usepackage{xcolor}

\def\BibTeX{{\rm B\kern-.05em{\sc i\kern-.025em b}\kern-.08em
    T\kern-.1667em\lower.7ex\hbox{E}\kern-.125emX}}

\usepackage{hyperref}

\usepackage[flushleft]{threeparttable}
\usepackage{tablefootnote}
\usepackage{array}

\newcolumntype{L}[1]{>{\raggedright\arraybackslash}p{#1}}
\usepackage{multicol}
\usepackage{multirow}
\usepackage{tabulary}
\usepackage{colortbl}

\usepackage{xcolor}
\usepackage{subcaption}
\usepackage{tabularx,booktabs}
\newcolumntype{L}[1]{>{\raggedright\arraybackslash}p{#1}}
\newcolumntype{C}[1]{>{\centering\arraybackslash}p{#1}}
\newcolumntype{R}[1]{>{\raggedleft\arraybackslash}p{#1}}
\usepackage[flushleft]{threeparttable}
\usepackage[ruled,vlined,german]{algorithm2e}
\usepackage{pdflscape}
\usepackage[switch,columnwise]{lineno}

\usepackage{url,hyperref,microtype}
\hypersetup{
    colorlinks=true,
    citecolor=blue,
    linkcolor=blue,
    filecolor=magenta,      
    urlcolor=black
}

\begin{document}

\title{Explainable graph attention network for stress recognition (StressGAT) via differential action units}

\author{\IEEEauthorblockN{Thomas Kassiotis}
\IEEEauthorblockA{\textit{Department of Electronic Engineering} \\
\textit{Hellenic Mediterranean University}\\
Chania, Greece \\
ddk305@edu.hmu.gr}

\and

\IEEEauthorblockN{Stefanos Gkikas}
\IEEEauthorblockA{\textit{Honda Research Institute Japan} \\
Wako City, Japan \\
stefanos.gkikas@jp.honda-ri.com}

\and

\IEEEauthorblockN{Nikolaos Smyrnis}
\IEEEauthorblockA{\textit{University Mental Health, Neurosciences and}\\
\textit{Precision Medicine, Research Institute ``COSTAS STEFANIS''} \\
\textit{2nd Department of Psychiatry, Medical School,} \\ 
\textit{National and Kapodistrian University of Athens} \\
Athens, Greece \\
smyrnis@med.uoa.gr}

\and

\IEEEauthorblockN{Giorgos Giannakakis}
\IEEEauthorblockA{\textit{Department of Electronic Engineering}\\ 
\textit{Hellenic Mediterranean University} \\
\textit{Institute of Computer Science,} \\
\textit{Foundation for Research and Technology Hellas} \\
Chania, Greece \\
ggian@hmu.gr}
}

\maketitle

\begin{abstract}

Stress is a dynamic process characterized by significant individual variability in facial expression. Traditional architectures, such as Recurrent Neural Networks (RNNs) and Convolutional Neural Networks (CNNs), often overlook person-specific baselines or lack the representational capacity to model the non-linear temporal progression of distress due to sequential bottlenecks and rigid grid-based constraints. Furthermore, many deep learning models lack the interpretability required for clinical deployment. This study introduces StressGAT, a Graph Attention Network that leverages the relational inductive bias of graph modeling to capture complex facial dynamics that indicate acute stress. By using Differential Action Units, the framework normalizes individual responses relative to neutral baselines to achieve personalized recognition. The proposed model achieves 88.62\% accuracy on a diverse stress-induction cohort (58 participants) using a subject-independent, Leave-One-Subject-Out (LOSO) cross-validation protocol. Beyond predictive accuracy, the architecture integrates a Multiple Instance Learning (MIL) attention mechanism to identify peak stress intervals and reveal distinct expressivity phenotypes. By simultaneously optimizing for accuracy and interpretability, this framework provides a robust, explainable solution for personalized affective monitoring.
\end{abstract}

\begin{IEEEkeywords}
stress detection, graph neural networks, facial action units, explainable AI (XAI), affective computing, multiple instance learning (MIL).
\end{IEEEkeywords}

\section{Introduction}
Stress estimation is a demanding task that evaluates responses using multiparametric behavioral and physiological data. While the majority of literature focuses on physiological biomarkers like cortisol or biosignals such as electrocardiogram (ECG) and electrodermal activity (EDA) \cite{8758154, GONZALEZCARABARIN2021106314, ZHOU2021106005}, there is emerging interest in Facial Expression Recognition (FER) as a non-invasive and scalable alternative \cite{Abdeldayem_Hamed_Nagy_2025, kumarihamy2026facial}. Although facial expressions can be manipulated, recent research identifies objective semi-voluntary parameters, such as micro-expressions and blink rates, as reliable stress indicators that reflect autonomic nervous system activation \cite{GIANNAKAKIS201789, DaudelinPeltier2017, https://doi.org/10.1155/2018/8734540}. These behavioral manifestations are tightly coupled to the autonomic arousal response, making them a promising modality for scalable, contact-free stress monitoring in unconstrained real-world environments.
 
A fundamental limitation of conventional approaches is the \textit{Temporal Accumulation Problem}. Acute stress is not a discrete, instantaneous event; rather, it reflects a non-stationary trajectory of allostatic load, in which the diagnostic signal is encoded in the temporal evolution of physiological and behavioral states, not in isolated snapshots. Traditional static architectures model facial activations as memoryless, independent events, lacking the inductive bias required to capture latent-state transitions, cumulative stress buildup, and the recovery dynamics that define the acute stress phenotype \cite{Chen_2026, 10204167, s21227498}. Recurrent models partially address this through sequential processing, yet their linear chain structure introduces information bottlenecks that limit the encoding of long-range, non-linear temporal dependencies critical to stress progression.
 
A second critical challenge is the \textit{Personalization Gap}, arising from significant inter-subject variability in baseline facial morphology \cite{10.3389/fpsyg.2026.1713462, Wang_2023, article}. Subject-agnostic architectures treat facial muscle activations as absolute measures and map them against population-level templates. This approach fails to disentangle permanent anatomical traits from transient, stress-induced deviations, leading to a distributional shift in which the neutral configuration of one individual may overlap with the stressed phenotype of another. Without subject-specific calibration, models remain susceptible to morphological noise, misinterpreting structural facial characteristics as affective signals and thereby inflating false-positive rates in cross-subject evaluations.

Beyond predictive performance, model transparency constitutes a critical requirement for deploying stress estimation systems in clinical and high-stakes environments \cite{badam2026explainable}. Domain experts require clear, physiologically grounded rationales for automated outputs to validate system decisions and ensure alignment with established psychological theory. Conventional deep learning architectures, while powerful, operate as black boxes that provide little insight into which facial regions or temporal intervals drive a given prediction \cite{alam2026embrace}. This opacity undermines clinical trust and hinders the adoption of automated affective monitoring tools in practice. In contrast, models that expose ranked feature contributions allow automated decisions to be traced back to specific physiological mechanisms \cite{kyprakis_gkikas_localization_2026}.
 
To address these limitations, this study introduces StressGAT, an explainable Graph Attention Network for personalized recognition of acute stress from facial behavior. The framework represents the stress response as a directed, causal, temporal graph, where GATv2 \cite{brody2022attentivegraphattentionnetworks} attention captures nonlinear stress buildup and recovery dynamics across sequential behavioral segments. Pairwise Differential Action Unit transformations normalize facial features relative to each subject's neutral baseline, thereby decoupling transient stress signals from fixed anatomical traits. Multiple Instance Learning (MIL) attentional pooling further enables the identification of peak stress intervals and provides anatomically grounded feature-level attribution. The proposed model achieves 88.62\% accuracy and an F1-score of 0.89 under a subject-independent Leave-One-Subject-Out (LOSO) protocol on a diverse cohort of 58 participants undergoing stress induction.

\section{Related Work}
While physiological biomarkers offer high-fidelity insights into Autonomic Nervous System (ANS) modulation, their real-world utility is frequently limited by compliance issues, motion artifacts, and the scalability constraints of skin-contact hardware \cite{Xu_Albeaino_2025, Hosseini2026, s25041241}. These limitations have motivated growing interest in vision-based stress estimation, where facial muscle activations and micro-expressive patterns serve as non-invasive, scalable proxies for autonomic arousal \cite{s20195552, valergaki2026combiningfacialvideosbiosignals, Walambe_2021}. Clinical research has established strong correlations between facial dynamics and physiological stress responses \cite{ZOU2022200, LEE201337}, supporting the viability of video-based affective monitoring in unconstrained environments.
 
Among vision-based representations, Facial Action Units (AUs), defined by the Facial Action Coding System (FACS) \cite{EKMAN1978}, have emerged as a robust alternative to holistic pixel-level features. AUs function as quantitative indices of localized muscle tension that remain resilient to head pose variations and illumination changes \cite{9320268}. Comparative evaluations demonstrate that AU-based representations generalize more effectively across diverse cohorts, as they decouple affective expression from subject identity, thereby mitigating the identity bias prevalent in black-box models \cite{liu2025actionunitenhancedynamic, Bouazizi2025}. However, AUs do not occur in isolation; they form a complex, interconnected topology in which spatial relationships among muscle groups carry significant diagnostic information that scalar representations fail to capture \cite{10669804, 10914864, designs9020045}.

Traditional sequential models, such as LSTMs and GRUs, are limited by linear chain structures that compress long-range dependencies into restricted hidden states, leading to information loss when modeling non-linear stress dynamics \cite{10.1108/JEIM-12-2020-0536, info15090517}. Grid-based convolutional architectures and vision transformers similarly lack spatial specificity, treating all pixel variations uniformly and allowing task-irrelevant motion to obscure subtle, localized affective signals \cite{xue2021transferlearningrelationawarefacial}. Evidence from adjacent facial assessment tasks indicates that partitioning the face into spatial regions prior to encoding recovers part of this lost specificity \cite{gkikas_reface_acii_2026}. These constraints have driven a shift toward Graph Neural Networks, which exploit the non-Euclidean topology of facial data to discover how localized activations propagate across regions and time.
 
Recent GNN-based approaches have demonstrated the value of relational modeling for affective tasks. Abedi and Khan \cite{abedi2024engagementmeasurementbasedfacial} applied a Spatial-Temporal Graph Convolutional Network to track facial landmark coordinates for engagement detection, though raw geometric shifts lack the muscular interpretability required to capture complex affective states. Hybrid CNN-GNN architectures \cite{Sarvakar2025} integrate convolutional feature extraction with graph-structured reasoning to capture spatial correlations within isolated frames, yet lack a temporal receptive field for modeling stress as a cumulative process. Bansal and Vyas \cite{article1} proposed a 3-layer GNN that exploits spatial dependencies among facial landmarks, but operating on absolute feature intensities from static images renders the model vulnerable to identity bias, where a subject's resting facial structure is misinterpreted as a stress response.
 
StressGAT addresses these limitations by transitioning from raw landmark coordinates to AU-based node representations encoding both sustained muscular contraction and temporal micro-expressive volatility. A directed causal temporal graph with GATv2 attention models the non-linear progression of stress across sequential behavioral segments, while pairwise differential normalization against subject-specific baselines suppresses morphological noise. MIL attentional pooling provides interpretable identification of peak stress intervals, grounding predictions in physiologically meaningful AU activations.

\section{Methodology}
\subsection{Facial Feature Extraction}
Action Unit intensities are extracted using the OpenFace 2.0 toolkit \cite{Baltruaitis2018OpenFace2F}, providing a standardized high-dimensional representation of facial behavior. OpenFace 2.0 is selected as the primary feature extractor due to its established robustness \cite{s21124222} in estimating AU intensities under unconstrained conditions, such as varying head poses and illumination changes, which are frequently encountered in stress-induction protocols. By utilizing Convolutional Experts Constrained Local Models (CE-CLM) \cite{zadeh2017convolutionalexpertsconstrainedlocal} for landmark detection, the toolkit ensures high spatial-temporal consistency, which is critical for the subsequent node representation. Table \ref{tab:aus_description} provides the anatomical and behavioral grounding for the extracted feature set, linking each investigated AU to its muscular basis and its established role in the facial manifestation of acute stress.

\begin{table}[ht]
\centering
\caption{Facial Action Units (AUs) and their Physiological Basis.}
\label{tab:aus_description}
\resizebox{\columnwidth}{!}{%
\begin{tabular}{@{}lll@{}}
\toprule
\textbf{AU} & \textbf{FACS Name} & \textbf{Muscular Basis} \\ \midrule
AU1 & Inner brow raiser & Frontalis (pars medialis) \\
AU2 & Outer brow raiser & Frontalis (pars lateralis) \\
AU4 & Brow lowerer & Depressor glabellae, d. supercilii, corrugator supercilii \\
AU5 & Upper lid raiser & Levator palpebrae superioris, superior tarsal muscle \\
AU6 & Cheek raiser & Orbicularis oculi (pars orbitalis) \\
AU7 & Lid tightener & Orbicularis oculi (pars palpebralis) \\
AU9 & Nose wrinkler & Levator labii superioris alaeque nasi \\
AU10 & Upper lip raiser & Levator labii superioris, caput infraorbitalis \\
AU12 & Lip corner puller & Zygomaticus major \\
AU14 & Dimpler & Buccinator \\
AU15 & Lip corner depressor & Depressor anguli oris (triangularis) \\
AU17 & Chin raiser & Mentalis \\
AU20 & Lip stretcher & Risorius \\
AU23 & Lip tightener & Orbicularis oris \\
AU25 & Lips part & Depressor labii inferioris, or relaxation of mentalis \\
AU26 & Jaw drop & Masseter; relaxed temporalis and internal pterygoid \\
AU45 & Blink & Relax. levator palpebrae, contr. orbicularis oculi \\ \bottomrule
\end{tabular}%
}
\end{table}


\subsection{Temporal Aggregation and Node Representation}
Frame-level AU intensities are aggregated into segment-level nodes using a 10-second non-overlapping sliding window. This duration optimizes the trade-off between macro-expression dynamics and temporal locality, a protocol consistent with established benchmarks in EngageNet \cite{singh2023iattentionlargescale}, DAiSEE \cite{gupta2022daiseeuserengagementrecognition}, and HBCU \cite{Whitehill2014TheFO}. Beyond its role as a standardized temporal anchor in cross-modal stress literature \cite{lall2025dynamicstressdetectionstudy, s21051678, article_1649691}, this 10-second interval was empirically validated through a sensitivity analysis (Table \ref{tab:node_sensitivity}). Experimental results demonstrate that this granularity outperforms both under-segmented ($N < 12$) and over-segmented ($N > 12$) configurations by maximizing contextual distinction. The use of disjoint temporal windows prevents data leakage between adjacent nodes, thereby ensuring the statistical independence of each segment during graph representation learning. For a video sequence sampled at 30 frames per second (FPS), each 10-second segment encapsulates $T = 300$ frames. Every frame $f$ within this interval is characterized by an Action Unit (AU) intensity vector $\mathbf{a}_f = [au_1, au_2, \dots, au_{17}]$, where $\mathbf{a}_f \in \mathbb{R}^{17}$. This extraction process yields a window-level observation matrix $\mathbf{M} \in \mathbb{R}^{T \times 17}$, which serves as the foundational data structure for subsequent node feature encoding. The pairwise transformation used to normalize against neutral baselines is based on the framework described in \cite{giannakakis2022automatic}.

The temporal observation matrix $\mathbf{M}$ is projected into a graph-ready node representation $\mathbf{x}_t \in \mathbb{R}^{34}$ by concatenating the arithmetic mean ($\mu$) and standard deviation ($\sigma$) of each Action Unit: $\mathbf{x}_t = [\mu_1, \dots, \mu_{17}]^\top \oplus [\sigma_1, \dots, \sigma_{17}]^\top$. This dual-moment strategy characterizes facial expressions via their temporal envelopes, effectively isolating sustained behaviors (high $\mu$, low $\sigma$) from transient dynamics (low $\mu$, high $\sigma$), which capture rapid micro-expressions indicative of emotional suppression \cite{10696279}. By distinguishing these patterns, the architecture filters stochastic noise while preserving salient affective signals for robust stress identification in the subsequent graph learning layers.

\begin{table}[ht]
\centering
\caption{Sensitivity Analysis of StressGAT Performance Across Varying Spatio-Temporal Granularities (Node Count)}
\label{tab:node_sensitivity}
\resizebox{\columnwidth}{!}{%
\begin{tabular}{lcccc}
\toprule
\textbf{Node Count ($N$)} & \textbf{Granularity State} & \textbf{Mean Acc. (\%)} & \textbf{Std. Dev.} & \textbf{Mean F1} \\ \midrule
2 Nodes  & Global Redundancy      & 79.12 \%          & $\pm$14.25          & 0.77          \\
5 Nodes  & High Overlap           & 80.45 \%          & $\pm$13.10          & 0.79          \\ \midrule
\textbf{12 Nodes} & \textbf{Optimal Balance} & \textbf{88.62 \%} & \textbf{$\pm$10.81} & \textbf{0.89} \\ \midrule
15 Nodes & Reduced Context        & 85.20 \%          & $\pm$11.45          & 0.84          \\
20 Nodes & Over-segmentation      & 84.13 \%          & $\pm$12.32          & 0.82          \\ \bottomrule
\end{tabular}
}
\vspace{4pt}
\begin{flushleft}
\footnotesize \textit{Note: The decline in performance beyond 12 nodes indicates that excessive temporal quantization leads to contextual fragmentation, hindering the attention mechanism's ability to aggregate meaningful physiological features.}
\end{flushleft}
\end{table}

\subsection{Causal Temporal Graph Construction}
To model the stress response as a cumulative and non-linear process, the aggregated node features are mapped onto a directed temporal graph $G = (\mathcal{V}, \mathcal{E})$, where $\mathcal{V} = \{v_1, v_2, \dots, v_N\}$ represents the sequence of $N=12$ temporal nodes. The architecture employs a causal adjacency strategy, in which edges are constructed so that a node at time $t$ receives information only from its preceding neighbors within a defined temporal receptive field. The edge set $\mathcal{E}$ is defined by the causality constraint:
\begin{equation}
(v_j, v_i) \in \mathcal{E} \iff 0 < i - j \leq \omega
\end{equation}
where $\omega$ denotes the look-back window. This directed connectivity ensures the hidden state of node $v_i$ is strictly conditioned on the historical context, effectively preventing look-ahead bias and mimicking the chronological progression of physiological stress. By enforcing a strictly lower-triangular adjacency matrix $\mathbf{A}$, the topology facilitates learning temporal dependencies via GAT attention coefficients while remaining viable for real-time inference. The resulting graph $G$ provides the foundation for message passing, enabling the model to aggregate multi-scale behavioral features across the entire video sequence.

\subsection{StressGAT Architecture}
\begin{figure*}[t] 
    \centering
    \includegraphics[width=\textwidth]{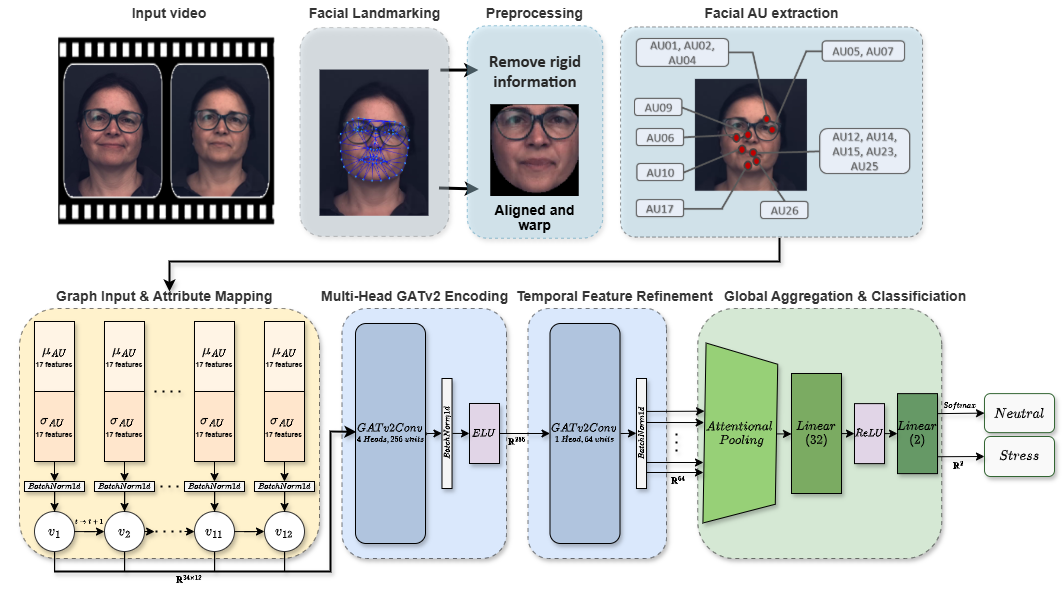} 
    \caption{System overview of the proposed StressGAT pipeline.} The framework consists of (i) facial feature extraction and Action Unit (AU) mapping; (ii) spatial-temporal graph construction with attribute normalization; (iii) multi-head GATv2 encoding for feature refinement; and (iv) attentional pooling for global feature aggregation and binary stress classification
    \label{fig:architecture}
\end{figure*}

The proposed StressGAT architecture, illustrated in Figure \ref{fig:architecture}, is designed to process temporal graph-structured data via a hierarchical feature-extraction pipeline. The specific layer configurations, input/output tensor dimensions, and computational complexity, totaling 56,935 trainable parameters, are detailed in Table \ref{tab:parameters}.

\begin{table}[ht]
\centering
\caption{Architectural Parameters of the StressGAT Model.}
\label{tab:parameters}
\resizebox{\columnwidth}{!}{
\begin{tabular}{@{}llccc@{}}
\toprule
\textbf{Stage} & \textbf{Layer (Type)} & \textbf{Input Shape} & \textbf{Output Shape} & \textbf{\# Param (M)} \\ \midrule
\multirow{2}{*}{\begin{tabular}[c]{@{}l@{}}Graph\\ Mapping\end{tabular}} 
    & BatchNorm1d & $[12, 34]$ & $[12, 34]$ & 68 \\
    & GATv2Conv (L1) & $[12, 34]$ & $[12, 256]$ & 18,432 \\ \midrule
\multirow{2}{*}{\begin{tabular}[c]{@{}l@{}}GNN\\ Backbone\end{tabular}} 
    & BatchNorm1d & $[12, 256]$ & $[12, 256]$ & 512 \\
    & GATv2Conv (L2) & $[12, 256]$ & $[12, 64]$ & 33,024 \\ \midrule
\multirow{4}{*}{\begin{tabular}[c]{@{}l@{}}Decision\\ Head\end{tabular}} 
    & BatchNorm1d & $[12, 64]$ & $[12, 64]$ & 128 \\
    & Attentional Pooling & $[12, 64]$ & $[1, 64]$ & 2,625 \\
    & Linear + ReLU & $[1, 64]$ & $[1, 32]$ & 2,080 \\
    & Linear (Output) & $[1, 32]$ & $[1, 2]$ & 66 \\ \midrule
\multicolumn{4}{@{}l}{\textbf{Total Trainable Parameters}} & \textbf{56,935} \\ \bottomrule
\end{tabular}%
}
\end{table}

\subsection{Dynamic Graph Backbone (GATv2)}
To stabilize feature distributions across the 17 Action Units, input nodes undergo Batch Normalization before processing. The backbone consists of two sequential GATv2 layers that use a dynamic attention mechanism to model non-linear transitions between vertices. Unlike static GATs, GATv2 enables adaptive neighbor ranking, which is critical for capturing fluctuating stress dynamics.

For a node $v_j$ and its causal predecessor $v_i$, the attention coefficient $e_{ij}$ is computed as:
\begin{equation}
e_{ij} = \vec{\mathbf{a}}^\top \text{LeakyReLU} \left( \mathbf{W} [h_i \parallel h_j] \right)
\end{equation}
where $\mathbf{W}$ is a shared weight matrix and $\vec{\mathbf{a}}$ is the learnable attention vector. These scores are normalized via softmax to yield weights $\alpha_{ij}$. The first layer employs four independent attention heads ($K=4$) followed by an ELU activation to project the initial $12 \times 34$ feature matrix into a 256-dimensional latent space. The second layer performs refinement, projecting these signals into a dense 64-dimensional representation $\mathcal{H} = \{h_1^{(2)}, \dots, h_{12}^{(2)}\}$ that captures localized temporal dependencies.

\subsection{Attentional Aggregation (MIL Pooling)}
To transition from individual vertex states to a unified graph-level embedding, the architecture utilizes Attentional Aggregation, a specialized implementation of Multiple Instance Learning (MIL). This approach acknowledges that stress-related facial behaviors may be transient and non-uniformly distributed across the total observation period. The pooling operation is controlled by a neural network gating mechanism—a Multi-Layer Perceptron (MLP)—that assigns a learnable importance weight to each temporal segment. The scalar score $s_i$ for each vertex embedding $h_i \in \mathbb{R}^{64}$ is calculated as:
\begin{equation}
s_i = \mathbf{w}^\top \rho(\mathbf{V}h_i + \mathbf{b})
\end{equation}
where $\mathbf{V} \in \mathbb{R}^{32 \times 64}$ and $\mathbf{w} \in \mathbb{R}^{32 \times 1}$ represent learnable weights, and $\rho(\cdot)$ denotes the ReLU activation. These scores are normalized via a softmax function to produce final attention weights $\alpha_i$, identifying the most salient temporal intervals (Peak Stress Nodes). The final graph-level representation $h_G$ is derived as the weighted summation of all vertex embeddings: 
\begin{equation}
h_G = \sum_{i=1}^{12} \alpha_i h_i
\end{equation}

\subsection{Classification and Optimization}
The final stage maps the encoded patterns to a definitive affective state via a Post-Pooling Decision Head. This head is a sequential MLP that reduces the dimensionality of $h_G$ to 32 units before projecting it into a 2-dimensional space representing the logits for Stress and Neutral classes. The final prediction $\hat{y}$ is obtained via a softmax function:
\begin{equation}
\hat{y} = \text{Softmax}(\mathbf{W}_2 \rho(\mathbf{W}_1 h_G + \mathbf{b}_1) + \mathbf{b}_2)
\end{equation}
The model is trained to minimize the Cross-Entropy Loss using the Adam optimizer with a learning rate of $10^{-3}$ and a weight decay of $10^{-4}$ to ensure robust generalization across the 58 subjects in the cohort. Generalization is further validated using a subject-independent Leave-One-Subject-Out (LOSO) protocol to prevent temporal data leakage and ensure person-agnostic stress detection

\section{Experimental Methodology}

\subsection{Experimental Protocol}
A comprehensive experimental protocol was developed to investigate facial and physiological responses under stress. The study incorporated neutral tasks to serve as a baseline, along with stress-inducing tasks that simulated and elicited stress through different stressor types. These stressors were categorized into four separate phases: \textit{social exposure}, \textit{emotional recall}, \textit{mental workload}, and \textit{stressful stimuli}. The specifics of the experimental tasks, including their durations and associated affective states, are detailed in Table \ref{tbl:experimental_tasks}.

\begin{table}[th]
    \small
	\centering
	\caption{Experimental tasks employed in this study}
	\begin{tabularx}
    {1\columnwidth}{C{0.1cm} C{0.1cm} L{4.6cm} C{0.9cm} C{0.9cm}}
    \toprule
		\textbf{\#}	& \textbf{}	& {\textbf{Experimental task}} & \textbf{Duration} & \textbf{Affective} \\
		\textbf{}	&    \textbf{}       &                 			   & \textbf{(min)}    & \textbf{State} \\
		\midrule
		\multicolumn{4}{c}{\textbf{Social exposure}}    \\
		1   &   1.1	& Neutral (reference)	 			&2 &N	\\
		2   &   1.2	& Baseline Description				&2 &N	\\
		3   &   1.3	& Interview		 					&2 &S	\\
		\hline
		\multicolumn{3}{c}{\textbf{Emotional recall}}   \\
		4   &   2.1	& Neutral (reference)				&2 &N	\\
		5   &   2.2	& Recall stressful event			   	&2 &S	\\
		\hline
		\multicolumn{3}{c}{\textbf{Mental workload}}   \\
    	6   &   3.1	& Reading letters/numbers (reference)   	&2 &N	\\
		7   &   3.2	& Stroop Colour-Word Test (SCWT)		&2 &S	\\
		8   &   3.3	& PASAT task         				&2 &S	\\
		\hline
		\multicolumn{3}{c}{\textbf{Stressful stimuli}}   \\
		9   &   4.1	& Relaxing video     				&2 &R  	\\
		10  &   4.2	& Adventure video					&2 &S  	\\
		11  &   4.3	& Psychological pressure video		&2 &S  	\\
		\bottomrule
		\multicolumn{4}{l}{\footnotesize Note: Intended affective state N:neutral, S:stress, R:relaxed)}  
        
	\end{tabularx}
	\vspace{0pt}
	\label{tbl:experimental_tasks}
    
\end{table}

\subsection{Stress induction protocol}
The \textit{social exposure} phase consisted of a brief interview in which participants described themselves, simulating the stress associated with public exposure.
During the \textit{emotional recall} phase, stress was induced by instructing participants to recall and mentally relive a personally stressful past event as though it were occurring in the present.
The \textit{mental tasks} phase evaluated cognitive load using two assessments: the modified Stroop Colour-Word Task (SCWT)~\cite{stroop1935studies}, in which participants read colour names printed in incongruent ink, with difficulty increased by alternating between reading and naming the ink colour; and the Paced Auditory Serial Addition Test (PASAT)~\cite{tombaugh2006comprehensive}, a neuropsychological assessment involving continuous mental arithmetic to measure attentional capacity under time constraints.
In the final \textit{stressful video} phase, participants viewed two-minute video clips intended to elicit emotional responses. The clips included both calming scenes and stress-inducing content, such as action sequences, scenarios involving heights for participants with mild acrophobia, home invasions, and car accidents.

\subsection{Study Dataset}
The experimental dataset comprised 58 adults (24 men, 34 women) with a mean age of 26.9±4.8 years. Each participant completed 11 tasks, including 4 neutral, 6 stressed, and 1 relaxed state. A neutral condition was administered at the beginning of each experimental phase and served as a baseline for subsequent stressful tasks. The inclusion criteria for this study were that participants be over 18 years old, and the exclusion criteria were a history of heart disease or failure to provide informed consent. All participants provided informed consent. The study received approval from the local Research Ethics Committee (approval no. 155/12-09-2022). The dataset is available (access is granted upon request for academic research purposes at the repository \footnote{\url{ https://github.com/ggian/stress_dataset}})\\

\section{Results and Discussion}
\subsection{Classification Perfomance}
To evaluate the advantages of relational graph modeling, the proposed architecture was benchmarked against baseline statistical classifiers, sequential recurrent architectures, and global self-attention mechanisms. As detailed in Table \ref{tab:final_architectures}, all deep learning architectures were constrained to a comparable model capacity to ensure that the reported performance gains are a result of structural efficiency.

The comparative classification results, summarized in Table \ref{tab:comparison_baselines}, demonstrate that StressGAT achieves the highest performance with a Mean Accuracy of 88.62\% and an F1-score of 0.89. StressGAT outperforms the Transformer (87.10\%) while requiring 12\% fewer trainable parameters. This efficiency gain suggests that the relational inductive bias, modeling 10-second temporal intervals as a structured directed graph, is more computationally efficient for stress recognition than the high-entropy complete-graph approach of global self-attention. By constraining the model's focus to specific temporal dependencies, the StressGAT minimizes over-parameterization and isolates the most salient features of the stress response, whereas the Transformer baseline likely suffers from the inherent noise of a fully connected attention matrix.

Furthermore, while the 1D-CNN baseline exhibited the highest stability ($\pm 9.20\%$), it lacks the interpretability of StressGAT's MIL Attention mechanism. Unlike recurrent baselines (LSTM/GRU), which rely on the last hidden state for classification, StressGAT uses hierarchical GATv2 layers to identify and weight specific stress-trigger intervals over the 120-second observation period.

\begin{table}[ht]
\centering
\caption{Comparative Summary of Model Architectures and Parametric Complexity.}
\label{tab:final_architectures}
\resizebox{\columnwidth}{!}{%
\begin{tabular}{@{}llccr@{}}
\toprule
\textbf{Model} & \textbf{Architectural Backbone} & \textbf{Heads} & \textbf{Aggregation} & \textbf{Params} \\ \midrule
LinearSVC      & Linear Hyperplane               & ---            & Feature Averaging    & $<$1k             \\
SVM (RBF)      & Non-linear Kernel               & ---            & Feature Averaging    & $<$1k             \\ \addlinespace[3pt]
LSTM     & 2-Layer LSTM                    & ---            & Last Hidden State    & 52k             \\
GRU      & 2-Layer GRU                     & ---            & Last Hidden State    & 41k             \\
1D-CNN         & Temporal Conv ($k=3$)           & ---            & Global Max Pool      & 28k             \\ \addlinespace[3pt]
Transformer    & Self-Attention                  & 4              & Global Avg Pool      & 65k             \\ \midrule
\textbf{StressGAT} & \textbf{Hierarchical GATv2} & \textbf{4 $\rightarrow$ 1} & \textbf{MIL Attention} & \textbf{57k} \\ \bottomrule
\end{tabular}%
}
\end{table}

\begin{table}[ht]
\centering
\caption{Benchmarking StressGAT against Traditional Machine Learning and State-of-the-Art Deep Learning Baselines}
\label{tab:comparison_baselines}
\resizebox{\columnwidth}{!}{
\begin{tabular}{lccc}
\toprule
\textbf{Architecture} & \textbf{Mean Acc. (\%)} & \textbf{Acc. Std. Dev} & \textbf{Mean F1} \\ \midrule
LinearSVM (Baseline)  & 69.21 \% & $\pm$15.26 & 0.68 \\
SVM (Baseline)        & 78.25 \% & $\pm$13.76 & 0.76 \\ \midrule
LSTM            & 82.15 \% & $\pm$12.10 & 0.80 \\
GRU             & 83.42 \% & $\pm$11.30 & 0.81 \\
StressGAT (Isolated Nodes) & 84.52 \% & $\pm$ 11.45 & 0.83 \\
1D-CNN                & 86.43 \% & $\pm$\textbf{9.20} & 0.85 \\ 
Transformer Self-Attention & 87.10 \% & $\pm$10.40 & 0.87 \\ \midrule
\textbf{StressGAT (Proposed)} & \textbf{88.62 \%} & $\pm$10.81 & \textbf{0.89} \\ \bottomrule
\end{tabular}
} 
\vspace{4pt}
\begin{flushleft}
\footnotesize 
\end{flushleft}
\end{table}
 The discriminative capability of the StressGAT model is further illustrated by the Receiver Operating Characteristic (ROC) curve presented in Figure \ref{fig:roc_curve}. Achieving an Area Under the Curve (AUC) of 0.907 signifies a high degree of class separability, indicating that the model effectively partitions the latent space between neutral and stress-induced states. 

 \begin{figure}[ht]
    \centering
    \label{fig:roc_curve}
    \includegraphics[width=0.8\linewidth]{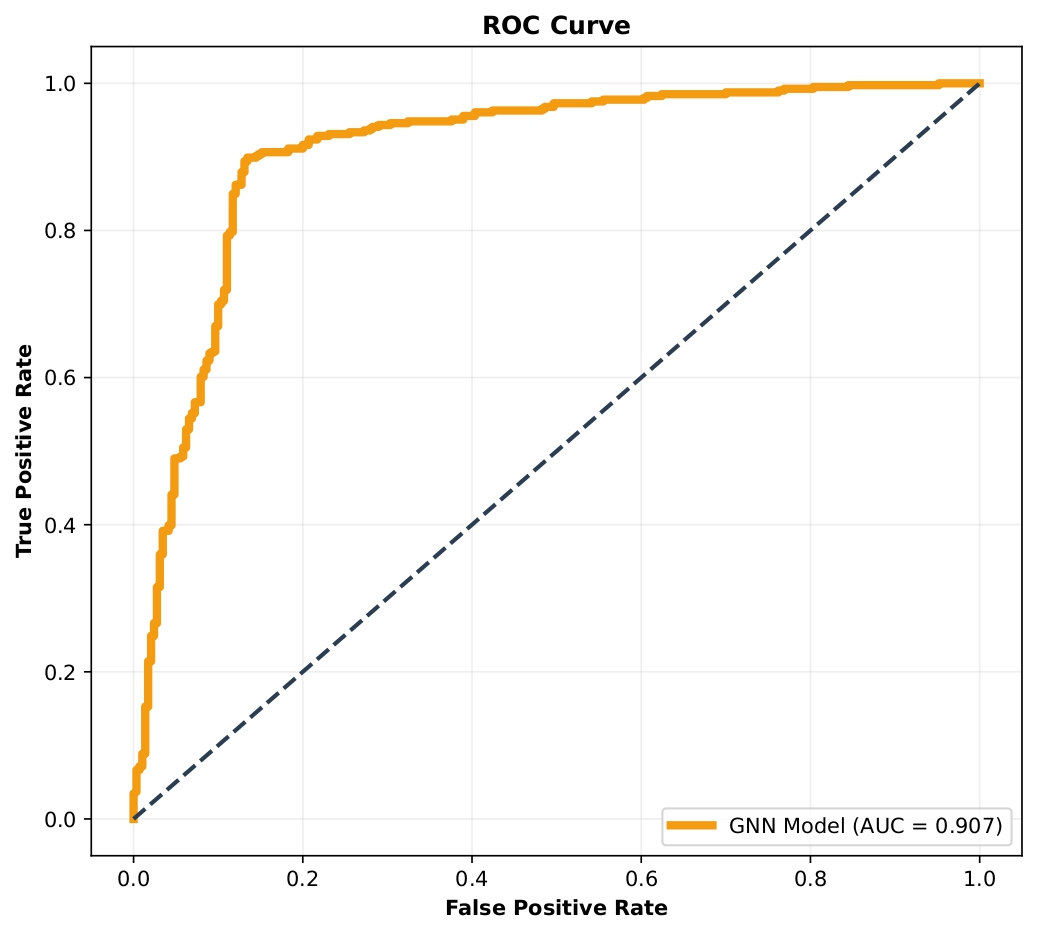}
    \caption{Receiver Operating Characteristic (ROC) Curve for the StressGAT model.}
    \label{fig:roc_curve}
\end{figure}

\subsection{Ablation Study: Impact of Temporal Topology}
An ablation study was implemented to evaluate the impact of temporal topology (StressGAT against Isolated Nodes). In this configuration, all temporal edges are removed, effectively reducing the graph to a series of disjointed static snapshots and forcing the model to rely exclusively on instantaneous spatial features. As demonstrated in Table~\ref{tab:comparison_baselines}, the inclusion of temporal connectivity yields a significant performance uplift. The removal of temporal edges results in a decrease in accuracy from 88.62\% to 84.52\%, with a corresponding $F_1$-score reduction from 0.89 to 0.83. These results validate the hypothesis that stress is more accurately characterized as a cumulative process rather than a collection of static, isolated events.

Analysis of the model’s misclassifications in Table \ref{tab:confusion_matrix} (39 False Positives, 35 False Negatives) highlights the inherent complexities of mapping subjective affect to objective labels. False negatives primarily occurred among extreme \textit{Suppressive} phenotypes, where autonomic arousal is decoupled from facial musculature. This near-complete expressive inhibition renders their stress kinematically indistinguishable from a neutral resting state, exposing a fundamental limitation of purely vision-based systems and underscoring the need for multimodal fusion, where physiological channels compensate for absent behavioral cues within a shared representation \cite{gkikas_workload_acii_2026, gkikas_arzate_pain_icmi_2026}. Conversely, false positives during baseline tasks were frequently driven by persistent morphological noise (e.g., heavy brow ridges mimicking AU4) or by ``evaluation apprehension,'' in which the experimental setup itself induced unlabelled, anticipatory anxiety. Thus, these errors largely reflect natural biological variance and subjective emotion regulation rather than algorithmic failure.


\begin{table}[htbp]
    \centering
    \caption{Confusion matrix of StressGAT performance. Diagonal elements denote correct classifications, while off-diagonal values represent Type I and Type II errors.}
    \label{tab:confusion_matrix}
    \renewcommand{\arraystretch}{1.4}
    \begin{tabular}{@{}llcc@{}}
        \toprule
        & & \multicolumn{2}{c}{\textbf{Predicted Label}} \\
        \cmidrule{3-4}
        & & \textbf{Neutral} & \textbf{Stress} \\
        \midrule
        \multirow{2}{*}{\textbf{True Label}} 
        & \textbf{Neutral} & 251 & 39 \\
        & \textbf{Stress}  & 35  & 323 \\
        \bottomrule
    \end{tabular}
    \label{tab:confusion_matrix}
\end{table}

\subsection{Phenotypical Analysis of Stress Response} 
\label{sct:phenotypical}
\begin{figure}[ht]
    \centering
    \includegraphics[width=1\linewidth]{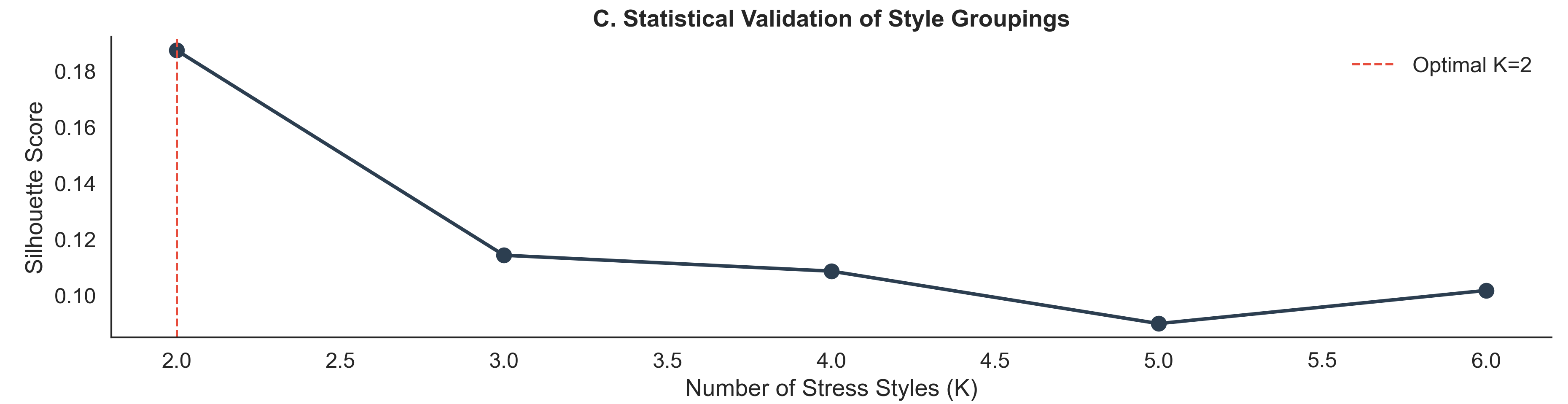}
    \caption{Clustering of participant stress phenotypes based on Facial Action Unit importance. The visual separation confirms the bifurcation into two distinct groups, as determined by the optimal silhouette score.}
    \label{fig:stress_groups}
\end{figure}

This section investigates the existence of distinct stress phenotypes based on facial expressive patterns. Utilizing a Leave-One-Subject-Out (LOSO) cross-validation scheme, the predictive importance of individual Action Units (AUs) was computed by the model for each participant across all experimental tasks. These personalized AU feature importance vectors were subsequently aggregated across all subjects to identify potential clusters representing distinct behavioral responses to stress. To determine the optimal number of clusters within this feature space, a Silhouette analysis Figure \ref{fig:stress_groups} was conducted. The resulting silhouette scores demonstrated a clear peak at two clusters, indicating the presence of two primary, quantifiable stress phenotypes among the participant cohort.

To characterize the underlying physiological nature of the identified stress responses, an in-depth evaluation of feature-importance distributions across both clusters was conducted. This analysis revealed a pronounced divergence in how psychological stress is mechanistically manifested, yielding two distinct participant categories: a predominantly populated \textbf{Expressive} (Externalizing) phenotype ($n=43$) and a \textbf{Suppressive} (Internalizing) phenotype ($n=15$). The defining Facial Action Units (AUs) and behavioral characteristics separating these two groups are summarized in Table \ref{tab:phenotype_comparison}.

\begin{table}[ht]
    \centering
    \footnotesize
    \caption{Summary of dominant Facial Action Units (AUs) and behavioral traits differentiating the Expressive and Suppressive stress phenotypes.}
    \label{tab:phenotype_comparison}
    \renewcommand{\arraystretch}{1.3}
    \begin{tabular}{@{} L{1.7cm} L{3.00cm} L{3.00cm} @{}}
        \toprule
        \textbf{Feature Category} & 
        \textbf{Phenotype 1: Expressive} \newline \textit{(Externalizing, $n=43$)} & 
        \textbf{Phenotype 2: Suppressive} \newline \textit{(Internalizing, $n=15$)} \\
        \midrule
        
        \textbf{Dominant Action Units} & 
        AU4 (Brow Lowerer), AU7 (Lid Tightener), AU9 (Nose Wrinkler), AU25/26 (Lips Part/Jaw Drop) & 
        AU14 (Dimpler), AU23 (Lip Tightener), AU24 (Lip Pressor), AU45 (Blink) \\
        
        \textbf{Anatomical \& Kinematic Profile} & 
        Upper face and open-mouth dynamics; high variance characterized by dynamic muscular contractions. & 
        Lower face (perioral) restriction; low variance characterized by static tension and autonomic blink rates. \\
        
        \textbf{Behavioral\newline Interpretation} & 
        Active, uninhibited manifestation of cognitive load, exasperation, or acute anxiety. & 
        Attempted emotion regulation, expressive restraint, and stoicism. \\
        \bottomrule
    \end{tabular}
\end{table}
The contrasting profiles demonstrate that the behavioral manifestation of stress is not universal, but rather governed by distinct coping strategies. The predominant Expressive phenotype is characterized by the uninhibited externalization of cognitive load, producing high-variance visual cues easily detectable by standard computer vision pipelines. In stark contrast, the Suppressive phenotype reveals a strategy of volitional emotion regulation and somatic containment. For this minority cohort, the psychological burden does not manifest as movement but as the absence of movement—specifically, through isometric lip-pressing and anomalous blink rates. The ability of the predictive framework to mathematically isolate these subtle, internalizing behaviors confirms that static facial rigidity and autonomic leakage are equally valid, measurable indicators of acute stress as dynamic facial expressions.

\section{Discussion}
This study introduced StressGAT, an explainable temporal Graph Attention Network designed for the personalized recognition of acute stress from facial behavior. By leveraging pairwise Differential Action Unit transformations and causal temporal graph modeling, the proposed architecture successfully addressed the personalization gap and temporal accumulation problems inherent in traditional static architectures. The proposed model, under subject-independent LOSO evaluation, achieved a classification accuracy of 88.62\%, outperforming state-of-the-art statistical classifiers (SVM), sequential architectures (LSTM, GRU), temporal convolutional models (1D-CNN), and global self-attention models (Transformer). 

The integration of Multiple Instance Learning (MIL) attentional pooling provided enhanced model transparency and explainability. MIL attentional pooling enabled the estimation of AU feature importance leading to two distinct stress pattern clusters based on the silhouette score: the \textit{Expressive} (externalizing) phenotype (n = 43) characterized by high-variance upper-face dynamics — AU4, AU7, AU9, and AU25/26 activations reflecting uninhibited externalization of arousal and the \textit{Suppressive} (internalizing) phenotype (n = 15) characterized by perioral restriction, isometric lip-pressing (AU23/24), and blink rate (AU45) reflecting volitional emotion regulation strategies that produce minimal overt movement as described in the section \ref{sct:phenotypical}. 

The present findings provide promising results for reliable, effective, and personalized stress recognition. Validating such systems under naturalistic, continuous-stress conditions, rather than discrete laboratory induction protocols, remains a challenge toward clinical translation. StressGAT, optimizing predictive accuracy and physiological interpretability, offers a highly scalable and transparent solution for affective monitoring.

\section*{Ethical Impact Statement}
Participants were recruited within a research center setting, mainly including graduate students and staff members. Eligibility required that individuals be over $18$ years old and provide written informed consent. Participants with a history of heart disease were excluded to maintain safety during stress-inducing procedures. Written informed consent was obtained from all participants prior to the experiment, and no personal information was recorded. The participants gave consent for their information to be used in the dataset. The dataset does not contain potentially offensive content. The study received approval from the Research Ethics Committee (approval no.~155/12-09-2022).
The proposed model aims to provide reliable stress assessment and interpretation. However, its application in real clinical settings requires further validation through well-designed clinical trials to verify robustness, generalizability, and safety before integration into practice.

The research or technology described in this study does not limit human rights. The research or technology can be used without deceiving people. The research or technology does not contain bias against certain groups of people that could result in discrimination.

\bibliographystyle{IEEEtran}
\bibliography{references}

@ARTICLE{8758154,
  author={Giannakakis, Giorgos and Grigoriadis, Dimitris and Giannakaki, Katerina and Simantiraki, Olympia and Roniotis, Alexandros and Tsiknakis, Manolis},
  journal={IEEE Transactions on Affective Computing}, 
  title={Review on Psychological Stress Detection Using Biosignals}, 
  year={2022},
  volume={13},
  number={1},
  pages={440-460},
  doi={10.1109/TAFFC.2019.2927350},
}

@article{GONZALEZCARABARIN2021106314,
title = {Machine Learning for personalised stress detection: Inter-individual variability of EEG-ECG markers for acute-stress response},
journal = {Computer Methods and Programs in Biomedicine},
volume = {209},
pages = {106314},
year = {2021},
issn = {0169-2607},
doi = {https://doi.org/10.1016/j.cmpb.2021.106314},
url = {},
author = {L. Gonzalez-Carabarin and E.A. Castellanos-Alvarado and P. Castro-Garcia and M.A. Garcia-Ramirez},
}

@article{ZHOU2021106005,
title = {ECG-based biometric under different psychological stress states},
journal = {Computer Methods and Programs in Biomedicine},
volume = {202},
pages = {106005},
year = {2021},
issn = {0169-2607},
doi = {https://doi.org/10.1016/j.cmpb.2021.106005},
url = {},
author = {Ruishi Zhou and Chenshuo Wang and Pengfei Zhang and Xianxiang Chen and Lidong Du and Peng Wang and Zhan Zhao and Mingyan Du and Zhen Fang}
}

@article{Abdeldayem_Hamed_Nagy_2025, 
title={Facial Expression Recognition: A Survey of Techniques, Datasets, and Real-World Challenges}, 
volume={15}, 
DOI={10.19139/soic-2310-5070-2789}, 
author={Abdeldayem, Mohamed and Hamed, Hesham F. A. and Nagy, Amr M.}, 
year={2025}, month={Oct.}, 
pages={733-761} 
}

@article{kumarihamy2026facial,
title={Facial Expression and Gesture Recognition System for Stress Detection with Deep Learning},
author={Kumarihamy, P. G. Dilini Kanchana},
journal={International Journal of Research and Scientific Innovation (IJRSI)},
volume={13},
number={2},
year={2026},
doi={10.51244/IJRSI.2026.130200165},
}

@article{GIANNAKAKIS201789,
title = {Stress and anxiety detection using facial cues from videos},
journal = {Biomedical Signal Processing and Control},
volume = {31},
pages = {89-101},
year = {2017},
issn = {1746-8094},
doi = {https://doi.org/10.1016/j.bspc.2016.06.020},
author = {G. Giannakakis and M. Pediaditis and D. Manousos and E. Kazantzaki and F. Chiarugi and P.G. Simos and K. Marias and M. Tsiknakis}
}

@article{DaudelinPeltier2017,
author = {Daudelin-Peltier, Camille and Forget, H{\'e}l{\`e}ne and Blais, Caroline and Desch{\^e}nes, Andr{\'e}a and Fiset, Daniel},
title = {The effect of acute social stress on the recognition of facial expression of emotions},
journal = {Scientific Reports},
year = {2017},
volume = {7},
number = {1},
pages = {1036},
month = {Apr},
day = {21},
doi = {10.1038/s41598-017-01053-3}
}

@article{https://doi.org/10.1155/2018/8734540,
author = {Bevilacqua, Fernando and Engström, Henrik and Backlund, Per},
title = {Automated Analysis of Facial Cues from Videos as a Potential Method for Differentiating Stress and Boredom of Players in Games},
journal = {International Journal of Computer Games Technology},
volume = {2018},
number = {1},
pages = {8734540},
doi = {https://doi.org/10.1155/2018/8734540},
year = {2018}
}

@article{Chen_2026,
title={Static for Dynamic: Towards a Deeper Understanding of Dynamic Facial Expressions Using Static Expression Data},
volume={17},
ISSN={2371-9850},
url={http://dx.doi.org/10.1109/TAFFC.2025.3623135},
DOI={10.1109/taffc.2025.3623135},
number={1},
journal={IEEE Transactions on Affective Computing},
publisher={Institute of Electrical and Electronics Engineers (IEEE)},
author={Chen, Yin and Li, Jia and Zhang, Yu and Hu, Zhenzhen and Shan, Shiguang and Wang, Meng and Hong, Richang},
year={2026},
month=jan, pages={438–451}
}

@INPROCEEDINGS{10204167,
author={Wang, Hanyang and Li, Bo and Wu, Shuang and Shen, Siyuan and Liu, Feng and Ding, Shouhong and Zhou, Aimin},
booktitle={2023 IEEE/CVF Conference on Computer Vision and Pattern Recognition (CVPR)}, 
title={Rethinking the Learning Paradigm for Dynamic Facial Expression Recognition}, 
year={2023},
pages={17958-17968},
doi={10.1109/CVPR52729.2023.01722}
}

@Article{s21227498,
AUTHOR = {Jeon, Taejae and Bae, Han Byeol and Lee, Yongju and Jang, Sungjun and Lee, Sangyoun},
TITLE = {Deep-Learning-Based Stress Recognition with Spatial-Temporal Facial Information},
JOURNAL = {Sensors},
VOLUME = {21},
YEAR = {2021},
NUMBER = {22},
ARTICLE-NUMBER = {7498},
PubMedID = {34833572},
ISSN = {1424-8220},
DOI = {10.3390/s21227498}
}

@ARTICLE{10.3389/fpsyg.2026.1713462,
AUTHOR={Shepelenko, Anna  and Kosonogov, Vladimir  and Shestakova, Anna N. },        
TITLE={Facial obstructions and baseline correction shape affective computing’s detection of emotion–behavior relationships},     
JOURNAL={Frontiers in Psychology},       
VOLUME={Volume 17 - 2026},
YEAR={2026},
DOI={10.3389/fpsyg.2026.1713462},
ISSN={1664-1078},
}

@article{Wang_2023,
title={Unlocking the Emotional World of Visual Media: An Overview of the Science, Research, and Impact of Understanding Emotion},
volume={111},
ISSN={1558-2256},
DOI={10.1109/jproc.2023.3273517},
number={10},
journal={Proceedings of the IEEE},
publisher={Institute of Electrical and Electronics Engineers (IEEE)},
author={Wang, James Z. and Zhao, Sicheng and Wu, Chenyan and Adams, Reginald B. and Newman, Michelle G. and Shafir, Tal and Tsachor, Rachelle},
year={2023},
month=oct, 
pages={1236–1286} 
}

@article{article,
author = {Tulsani, Vijya},
year = {2025},
month = {04},
pages = {427-436},
title = {Real-Time Facial Emotion Recognition for AI-Enhanced Personalized Learning},
volume = {10},
journal = {Journal of Information Systems Engineering and Management},
doi = {10.52783/jisem.v10i40s.7313}
}

@article{alam2026embrace,
  title={Explainable Multitask Burnout Prediction Using Adaptive Deep Learning (EMBRACE) for Resident Physicians: Algorithm Development and Validation Study},
  author={Alam, S and Alam, MAU},
  journal={JMIR AI},
  volume={5},
  pages={e57025},
  year={2026},
  month={Jan},
  publisher={JMIR Publications Inc., Toronto, Canada},
  doi={10.2196/57025}
}

@article{badam2026explainable,
  title={Explainable AI in High-Stakes Domains: Improving Trust, Transparency, And Accountability in Automated Decision-Making},
  author={Badam, S.},
  journal={European Journal of Computer Science and Information Technology},
  volume={14},
  number={2},
  pages={13--34},
  year={2026}
}

@misc{brody2022attentivegraphattentionnetworks,
      title={How Attentive are Graph Attention Networks?}, 
      author={Shaked Brody and Uri Alon and Eran Yahav},
      year={2022},
      eprint={2105.14491},
      archivePrefix={arXiv},
      primaryClass={cs.LG},
      url={https://arxiv.org/abs/2105.14491}, 
}

@article{Hosseini2026,
author = {Hosseini, M. and Gottumukkala, R. and Bhupatiraju, R. and Maida, A. and Chu, H.},
title = {Wearable-Based Stress Detection for Real-World Data: Perspective on Challenges and Recommendations},
journal = {JMIR Preprints},
year = {2026},
month = {February},
day = {18},
note = {Preprint ID: 93741},
}

@Article{s25041241,
AUTHOR = {Alkurdi, Abdulrahman and He, Maxine and Cerna, Jonathan and Clore, Jean and Sowers, Richard and Hsiao-Wecksler, Elizabeth T. and Hernandez, Manuel E.},
TITLE = {Extending Anxiety Detection from Multimodal Wearables in Controlled Conditions to Real-World Environments},
JOURNAL = {Sensors},
VOLUME = {25},
YEAR = {2025},
NUMBER = {4},
ARTICLE-NUMBER = {1241},
PubMedID = {40006470},
ISSN = {1424-8220},
DOI = {10.3390/s25041241}
}

@article{Xu_Albeaino_2025,
title={A systematic and bibliometric review on physiological monitoring systems and wearable sensing devices for mental status monitoring in construction: Trends, limitations, and future directions},
author={Xu, D. and Albeaino, G.},
journal={Journal of Information Technology in Construction (ITcon)},
volume={30},
pages={1814--1865},
year={2025},
publisher={International Council for Research and Innovation in Building and Construction (CIB)},
doi={10.36680/j.itcon.2025.075}
}

@article{Walambe_2021,
title={Employing Multimodal Machine Learning for Stress Detection},
volume={2021},
ISSN={2040-2295},
DOI={10.1155/2021/9356452},
journal={Journal of Healthcare Engineering},
publisher={Wiley},
author={Walambe, Rahee and Nayak, Pranav and Bhardwaj, Ashmit and Kotecha, Ketan},
editor={Jain, Deepak Kumar},
year={2021},
month=oct, pages={1–12} 
}

@Article{s20195552,
AUTHOR = {Zhang, Huijun and Feng, Ling and Li, Ningyun and Jin, Zhanyu and Cao, Lei},
TITLE = {Video-Based Stress Detection through Deep Learning},
JOURNAL = {Sensors},
VOLUME = {20},
YEAR = {2020},
NUMBER = {19},
ARTICLE-NUMBER = {5552},
PubMedID = {32998327},
ISSN = {1424-8220},
DOI = {10.3390/s20195552}
}

@misc{valergaki2026combiningfacialvideosbiosignals,
title={Combining Facial Videos and Biosignals for Stress Estimation During Driving}, 
author={Paraskevi Valergaki and Vassilis C. Nicodemou and Iason Oikonomidis and Antonis Argyros and Anastasios Roussos},
year={2026},
eprint={2601.04376},
archivePrefix={arXiv},
primaryClass={cs.CV},
url={https://arxiv.org/abs/2601.04376}, 
}

@article{LEE201337,
title = {An amplification of feedback from facial muscles strengthened sympathetic activations to emotional facial cues},
journal = {Autonomic Neuroscience},
volume = {179},
number = {1},
pages = {37-42},
year = {2013},
issn = {1566-0702},
doi = {https://doi.org/10.1016/j.autneu.2013.06.009},
author = {In-Seon Lee and Sung-Soo Yoon and Soon-Ho Lee and Hyejung Lee and Hi-Joon Park and Christian Wallraven and Younbyoung Chae}
}

@article{ZOU2022200,
title = {Concordance between facial micro-expressions and physiological signals under emotion elicitation},
journal = {Pattern Recognition Letters},
volume = {164},
pages = {200-209},
year = {2022},
issn = {0167-8655},
doi = {https://doi.org/10.1016/j.patrec.2022.11.001},
author = {Bochao Zou and Yingxue Wang and Xiaolong Zhang and Xiangwen Lyu and Huimin Ma},
}

@ARTICLE{10669804,
  author={Kyrou, Maria and Kompatsiaris, Ioannis and Petrantonakis, Panagiotis C.},
  journal={IEEE Transactions on Affective Computing}, 
  title={Deep Learning Approaches for Stress Detection: A Survey}, 
  year={2025},
  volume={16},
  number={2},
  pages={499-517},
  doi={10.1109/TAFFC.2024.3455371}}

@INPROCEEDINGS{10914864,
  author={M, Sindu and Ashwin, T and V, Shyam Sunder and S, Shivasankaran and P, Poongothai and Ponnusamy, R.},
  booktitle={2024 IEEE International Women in Engineering (WIE) Conference on Electrical and Computer Engineering (WIECON-ECE)}, 
  title={Real-Time Stress Detection Using Facial Images and Convolutional Neural Networks}, 
  year={2024},
  pages={432-437},
  doi={10.1109/WIECON-ECE64149.2024.10914864}
  }

@Article{designs9020045,
AUTHOR = {Uddin, Jia},
TITLE = {A Novel Lightweight Deep Learning Approach for Drivers’ Facial Expression Detection},
JOURNAL = {Designs},
VOLUME = {9},
YEAR = {2025},
NUMBER = {2},
ARTICLE-NUMBER = {45},
ISSN = {2411-9660},
DOI = {10.3390/designs9020045}
}

@book{Ekman1978,
author = {Ekman, Paul and Friesen, Wallace V.},
title = {Facial Action Coding System: A Technique for the Measurement of Facial Movement},
publisher = {Consulting Psychologists Press},
address = {Palo Alto, CA},
year = {1978},
}

@INPROCEEDINGS{9320268,
author={Giannakakis, Giorgos and Koujan, Mohammad Rami and Roussos, Anastasios and Marias, Kostas},
booktitle={2020 15th IEEE International Conference on Automatic Face and Gesture Recognition (FG 2020)}, 
title={Automatic stress detection evaluating models of facial action units}, 
year={2020},
volume={},
number={},
pages={728-733},
doi={10.1109/FG47880.2020.00129}
}

@article{Bouazizi2025,
author = {Bouazizi, Mondher and Feghoul, Kevin and Wang, Shengze and Yin, Yue and Ohtsuki, Tomoaki},
title = {A Non-Invasive Approach for Facial Action Unit Extraction and Its Application in Pain Detection},
journal = {Bioengineering},
year = {2025},
volume = {12},
number = {2},
pages = {195},
doi = {10.3390/bioengineering12020195},
pmid = {40001714},
pmcid = {PMC11851526}
}

@misc{liu2025actionunitenhancedynamic,
title={Action Unit Enhance Dynamic Facial Expression Recognition}, 
author={Feng Liu and Lingna Gu and Chen Shi and Xiaolan Fu},
year={2025},
eprint={2507.07678},
archivePrefix={arXiv},
primaryClass={cs.CV},
url={https://arxiv.org/abs/2507.07678}, 
}

@article{10.1108/JEIM-12-2020-0536,
author = {Kumar, Saurabh},
title = {Deep learning based affective computing},
journal = {Journal of Enterprise Information Management},
volume = {34},
number = {5},
pages = {1551-1575},
year = {2021},
month = {10},
issn = {1741-0398},
doi = {10.1108/JEIM-12-2020-0536},
}

@Article{info15090517,
AUTHOR = {Mienye, Ibomoiye Domor and Swart, Theo G. and Obaido, George},
TITLE = {Recurrent Neural Networks: A Comprehensive Review of Architectures, Variants, and Applications},
JOURNAL = {Information},
VOLUME = {15},
YEAR = {2024},
NUMBER = {9},
ARTICLE-NUMBER = {517},
ISSN = {2078-2489},
DOI = {10.3390/info15090517}
}

@misc{xue2021transferlearningrelationawarefacial,
title={TransFER: Learning Relation-aware Facial Expression Representations with Transformers}, 
author={Fanglei Xue and Qiangchang Wang and Guodong Guo},
year={2021},
eprint={2108.11116},
archivePrefix={arXiv},
primaryClass={cs.CV},
url={https://arxiv.org/abs/2108.11116}, 
}

@misc{abedi2024engagementmeasurementbasedfacial,
title={Engagement Measurement Based on Facial Landmarks and Spatial-Temporal Graph Convolutional Networks}, 
author={Ali Abedi and Shehroz S. Khan},
year={2024},
eprint={2403.17175},
archivePrefix={arXiv},
primaryClass={cs.CV},
url={https://arxiv.org/abs/2403.17175}, 
}

@article{Sarvakar2025,
  author = {Sarvakar, K. and Rana, K.},
  title = {Revolutionizing facial emotion recognition: in-depth analysis of cutting-edge models, methodologies, and datasets},
  journal = {Discovery Artificial Intelligence},
  year = {2025},
  volume = {5},
  number = {388},
  doi = {10.1007/s44163-025-00553-w}
}

@article{article1,
author = {Bansal, Megha and Vyas, Vaibhav},
year = {2024},
month = {03},
pages = {96-102},
title = {An evolutionary and insightful graph neural network based hybrid model for deciphering tenacious stress detection of humans using facial emotion recognition: coinage to affective computing},
volume = {17},
journal = {SciEnggJ},
doi = {10.54645/202417SupZAM-51}
}

@inproceedings{gkikas_reface_acii_2026,
  title={{ReFace: Reorganizing Facial Spatiotemporal Representations for Improved Pain Assessment}},
  author={Stefanos Gkikas and Yu Fang and Christian Arzate Cruz and Muhammad Umar Khan and Raul Fernandez Rojas},
  booktitle={2026 14th International Conference on Affective Computing and Intelligent Interaction (ACII)},
  year={2026},
  location={Puebla, Mexico},
  publisher={IEEE}
}

@inproceedings{gkikas_workload_acii_2026,
  title={{Towards a Unified Modality-Agnostic Multimodal Framework for Cognitive Workload Assessment}},
  author={Stefanos Gkikas and Christian Arzate Cruz and Calvin Joseph and Giorgos Giannakakis and Raul Fernandez Rojas},
  booktitle={2026 14th International Conference on Affective Computing and Intelligent Interaction (ACII)},
  year={2026},
  location={Puebla, Mexico},
  publisher={IEEE}
}

@inproceedings{gkikas_arzate_pain_icmi_2026,
  title={{A Unified Tokenization Framework for Pain Recognition using Heterogeneous 3D Modalities}},
  author={Stefanos Gkikas and Christian Arzate Cruz and Valentina Becchetti and Muhammad Umar Khan and Alessandro Giuseppi and Raul Fernandez Rojas},
  booktitle={Proceedings of the 28th ACM International Conference on Multimodal Interaction},
  year={2026},
  location={Napoli, Italy},
  publisher={Association for Computing Machinery}
}

@misc{kyprakis_gkikas_localization_2026,
  title         = {An Exploratory Analysis of Pain Localization via Explainable Computational Modeling},
  author        = {Kyprakis, Ioannis and Gkikas, Stefanos and Nichols, Eric and Fang, Yu and Tsiknakis, Manolis},
  year          = {2026},
  eprint        = {2601.00000},
  archivePrefix = {arXiv},
  primaryClass  = {cs.LG}
}

@article{Baltruaitis2018OpenFace2F,
title={OpenFace 2.0: Facial Behavior Analysis Toolkit},
author={Tadas Baltru{\v{s}}aitis and Amir Zadeh and Yao Chong Lim and Louis-philippe Morency},
journal={2018 13th IEEE International Conference on Automatic Face \& Gesture Recognition (FG 2018)},
year={2018},
pages={59-66},
}

@Article{s21124222,
AUTHOR = {Namba, Shushi and Sato, Wataru and Osumi, Masaki and Shimokawa, Koh},
TITLE = {Assessing Automated Facial Action Unit Detection Systems for Analyzing Cross-Domain Facial Expression Databases},
JOURNAL = {Sensors},
VOLUME = {21},
YEAR = {2021},
NUMBER = {12},
ARTICLE-NUMBER = {4222},
PubMedID = {34203007},
ISSN = {1424-8220},
DOI = {10.3390/s21124222}
}

@misc{zadeh2017convolutionalexpertsconstrainedlocal,
title={Convolutional Experts Constrained Local Model for Facial Landmark Detection}, 
author={Amir Zadeh and Tadas Baltru\v{s}aitis and Louis-Philippe Morency},
year={2017},
eprint={1611.08657},
archivePrefix={arXiv},
primaryClass={cs.CV},
url={https://arxiv.org/abs/1611.08657}, 
}

@misc{singh2023iattentionlargescale,
title={Do I Have Your Attention: A Large Scale Engagement Prediction Dataset and Baselines}, 
author={Monisha Singh and Ximi Hoque and Donghuo Zeng and Yanan Wang and Kazushi Ikeda and Abhinav Dhall},
year={2023},
eprint={2302.00431},
archivePrefix={arXiv},
primaryClass={cs.CV},
url={https://arxiv.org/abs/2302.00431}, 
}

@misc{gupta2022daiseeuserengagementrecognition,
title={DAiSEE: Towards User Engagement Recognition in the Wild}, 
author={Abhay Gupta and Arjun D'Cunha and Kamal Awasthi and Vineeth Balasubramanian},
year={2022},
eprint={1609.01885},
archivePrefix={arXiv},
primaryClass={cs.CV},
url={https://arxiv.org/abs/1609.01885}, 
}

@article{Whitehill2014TheFO,
title={The Faces of Engagement: Automatic Recognition of Student Engagement from Facial Expressions},
author={Jacob Whitehill and Zewelanji N. Serpell and Yi-Ching Lin and Aysha Foster and Javier R. Movellan},
journal={IEEE Transactions on Affective Computing},
year={2014},
volume={5},
pages={86-98}
}

@misc{lall2025dynamicstressdetectionstudy,
title={Dynamic Stress Detection: A Study of Temporal Progression Modelling of Stress in Speech}, 
author={Vishakha Lall and Yisi Liu},
year={2025},
eprint={2510.08586},
archivePrefix={arXiv},
primaryClass={eess.AS},
}

@Article{s21051678,
AUTHOR = {Keelawat, Panayu and Thammasan, Nattapong and Numao, Masayuki and Kijsirikul, Boonserm},
TITLE = {A Comparative Study of Window Size and Channel Arrangement on EEG-Emotion Recognition Using Deep CNN},
JOURNAL = {Sensors},
VOLUME = {21},
YEAR = {2021},
NUMBER = {5},
ARTICLE-NUMBER = {1678},
PubMedID = {33804366},
ISSN = {1424-8220},
DOI = {10.3390/s21051678}
}

@article{article_1649691, 
title={DEVELOPMENT OF A TERNARY LEVELS EMOTION CLASSIFICATION MODEL UTILIZING ELECTROENCEPHALOGRAPHY DATA SET}, journal={Konya Journal of Engineering Sciences}, 
volume={13}, 
pages={607–623}, 
year={2025}, 
DOI={10.36306/konjes.1649691}, 
author={Okumuş, Hatice and Ergün, Ebru},
number={2} 
}

@INPROCEEDINGS{10696279,
author={Gupta, Pooja and Maji, Srabanti and Jain, Vijay Kumar and Agarwal, Shipra},
booktitle={2024 First International Conference on Pioneering Developments in Computer Science \& Digital Technologies (IC2SDT)}, 
title={Automatic Stress Recognition Using FACS from Prominent Facial Regions}, 
year={2024},
pages={488-492},
doi={10.1109/IC2SDT62152.2024.10696279}
}

@article{stroop1935studies,
  title={Studies of interference in serial verbal reactions},
  author={Stroop, J. Ridley},
  journal={Journal of Experimental Psychology},
  volume={18},
  number={6},
  pages={643--662},
  year={1935},
  publisher={Psychological Review Company}
}

@article{tombaugh2006comprehensive,
  title={A comprehensive review of the Paced Auditory Serial Addition Test (PASAT)},
  author={Tombaugh, T. N.},
  journal={Archives of Clinical Neuropsychology},
  volume={21},
  number={1},
  pages={53--76},
  year={2006},
  publisher={Oxford University Press},
  doi={10.1016/j.acn.2005.07.006}
}

@article{giannakakis2022automatic,
  title={Automatic stress analysis from facial videos based on deep facial action units recognition},
  author={Giannakakis, Giorgos and Koujan, Mohammad Rami and Roussos, Anastasios and Marias, Kostas},
  journal={Pattern analysis and applications},
  volume={25},
  number={3},
  pages={521--535},
  year={2022},
  publisher={Springer}
}

\end{document}